%% file: template.tex
\documentclass[journal]{vgtc}                     
\graphicspath{{figures/}{pictures/}{images/}{./}} 

\PassOptionsToPackage{svgnames}{xcolor}
\usepackage{xcolor}

\usepackage{microtype}                 
\PassOptionsToPackage{warn}{textcomp}  
\usepackage{textcomp}                  
\usepackage{mathptmx}                  
\usepackage{times}                     
\usepackage{cite}                      
\usepackage{tabu}                      
\usepackage{booktabs}                  
\usepackage{subfig}
\usepackage{svg}
\renewcommand\footnotemark{}

\usepackage{multirow}

\usepackage{float}

\usepackage{tabularx}

\usepackage{colortbl}
\usepackage{diagbox}  
\usepackage{booktabs} 
\usepackage{amssymb}

\newcommand{\rb}[1]{{\color{black} #1}}

\onlineid{0}

\vgtccategory{Research}

\usepackage{amsmath}
\usepackage{mdframed}
\usepackage{makecell}
\usepackage{url}
\usepackage[most]{tcolorbox}
\usepackage{enumitem}

\title{Detecting Visual Information Manipulation Attacks in Augmented Reality: A Multimodal Semantic Reasoning Approach}

\author{%
  Yanming Xiu,
  Maria Gorlatova
}

\authorfooter{
  \item
  	Yanming Xiu is with Duke University.
  	E-mail: yanming.xiu@duke.edu

  \item 
        Maria Gorlatova is with Duke University.
        E-mail: maria.gorlatova@duke.edu
  
}

\abstract{
    The virtual content in augmented reality (AR) can introduce misleading or harmful information, leading to semantic misunderstandings or user errors. In this work, we focus on visual information manipulation (VIM) attacks in AR, where virtual content changes the meaning of real-world scenes in subtle but impactful ways. We introduce a taxonomy that categorizes these attacks into three formats: character, phrase, and pattern manipulation, and three purposes: information replacement, information obfuscation, and extra wrong information. Based on the taxonomy, we construct a dataset, AR-VIM, which consists of \rb{452} raw-AR video pairs spanning \rb{202} different scenes, each simulating a real-world AR scenario. To detect the attacks in the dataset, we propose a multimodal semantic reasoning framework, VIM-Sense. It combines the language and visual understanding capabilities of vision-language models (VLMs) with optical character recognition (OCR)-based textual analysis. VIM-Sense achieves an attack detection accuracy of \rb{88.94}\% on AR-VIM, consistently outperforming vision-only and text-only baselines. The system achieves an average attack detection latency of \rb{7.07} seconds in a simulated video processing framework and 7.17 seconds in a real-world evaluation conducted on a mobile Android AR application.
} 

\keywords{Mixed / Augmented Reality, Vision Language Models, Scene Understanding, Information Manipulation, Optical Character Recognition}

\graphicspath{{figs/}{figures/}{pictures/}{images/}{./}} 

\usepackage{tabu}                      
\usepackage{booktabs}                  
\usepackage{lipsum}                    
\usepackage{mwe}                       

\usepackage{mathptmx}                  

\renewcommand\footnotemark{}

\usepackage{multirow}

\usepackage{float}

\usepackage{tabularx}

\usepackage{colortbl}

\onlineid{1307}

\vgtccategory{Research}

\usepackage{amsmath}

\usepackage{mdframed}
\usepackage{makecell}
\usepackage{url}

\begin{document}



\firstsection{Introduction}

\maketitle
\input{sections/1_introduction}

\input{sections/2_related_work}
\input{sections/3_VIMA_modeling}
\input{sections/4_AR_dataset}

\input{sections/5_system_design}
\input{sections/6_evaluation}
\input{sections/7_limitation}
\input{sections/8_conclusion}

\acknowledgments{
We thank the participants of our user-based label validation for their invaluable effort and assistance in this research. This work was supported in part by NSF grants CSR-2312760, CNS-2112562, and IIS-2231975, NSF CAREER Award IIS-2046072, NSF NAIAD Award 2332744, a Cisco Research Award, a Meta Research Award, Defense Advanced Research Projects Agency Young Faculty Award HR0011-24-1-0001, and the Army Research Laboratory under Cooperative Agreement Number W911NF-23-2-0224. The views and conclusions contained in this document are those of the authors and should not be interpreted as representing the official policies, either expressed or implied, of the Defense Advanced Research Projects Agency, the Army Research Laboratory, or the U.S. Government. This paper has been approved for public release; distribution is unlimited. No official endorsement should be inferred. The U.S.~Government is authorized to reproduce and distribute reprints for Government purposes notwithstanding any copyright notation herein.
}

\bibliographystyle{abbrv-doi}

\bibliography{template}








\end{document}

%% file: sections/1_introduction.tex

\begin{figure}[t]
\includegraphics[width=0.98\linewidth]{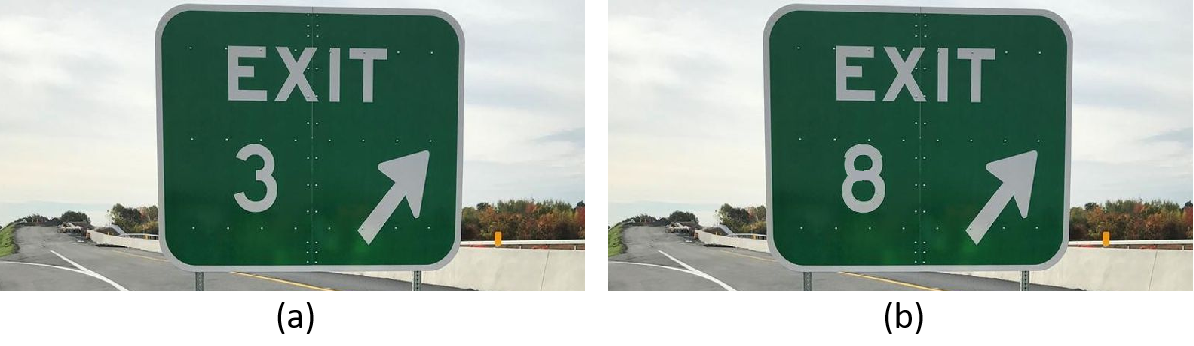}
\centering
\vspace{-0.3cm}
\caption{An example of a visual information manipulation attack in AR. (a): The real-world view of a highway exit sign indicating "Exit 3". (b): The AR view, where virtual content (a horizontally flipped "3") has been added to subtly change the number from "3" to "8".}
\label{fig:vim example}
\vspace{-0.5cm}
\end{figure}

Augmented Reality (AR) overlays virtual content onto the real world to enhance user perception, improve decision-making, and enrich interactive experiences. While AR offers benefits across domains such as education, navigation, and entertainment, it also introduces new risks when virtual content is poorly designed or intentionally misleading, creating visual attacks in AR\cite{attack01, attack02,attack03, misleading01, viddar}. In particular, visual information manipulation attacks, in which virtual content changes the meaning or interpretation of real-world scenes, can lead to serious consequences, such as misunderstanding warnings, misidentifying important objects, or taking action based on incorrect, manipulated information\cite{VIM01, VIM02, VIM03}. For example, in \cref{fig:vim example}, the virtual object (the horizontally flipped "3") is overlaid on a real-world object, subtly altering the exit number from "3" to "8." While visually similar, the change has significant semantic implications and may mislead users during navigation. These attacks are particularly critical because they are often subtle, visually realistic, and seamlessly integrated into the scene, making them difficult for users to notice and for automated evaluation tools to detect.

\begin{figure}[t]
\includegraphics[width=0.98\linewidth]{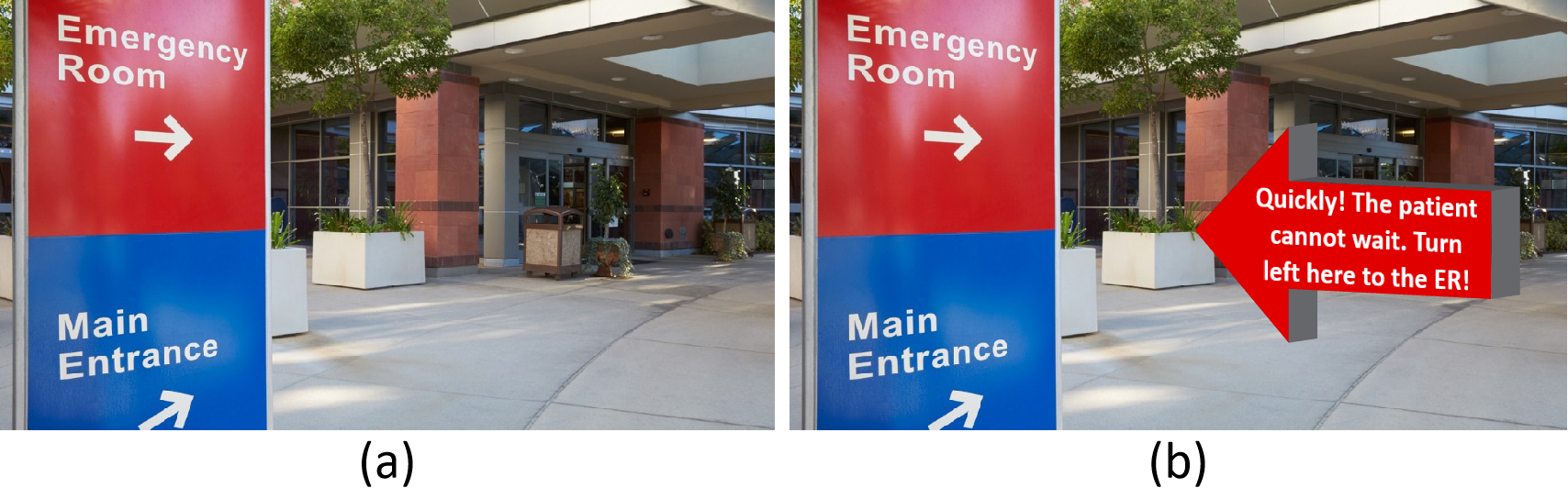}
\centering
\vspace{-0.3cm}
\caption{An example of wrong information added in an AR scene. (a) The real-world view shows that the Emergency Room is to the right. (b) The AR view introduces a virtual arrow that falsely instructs users to turn left. Although the original content is untouched, the meaning of the scene is changed and could mislead users in urgent situations.}
\label{fig:vim example 2}
\vspace{-0.5cm}
\end{figure}

Detecting visual information manipulation attacks requires reasoning about both the virtual content and how it interacts with the real-world scene. Prior work on AR content quality assessment has primarily focused on evaluating visual properties of virtual elements, such as shadow consistency\cite{ARassessshadow01}, lighting effect\cite{ARassessmentlighting01}, and spatial alignment\cite{ARassessphysics01, ARassessmisalign01, ARassessmisalign02, ARassessmisalign03} to determine how well the virtual content blends into the environment, or whether it affects users' perception\cite{ARassess01, ARassesslighting02, ARIQA01}. While important for visual coherence, these approaches fall short when it comes to semantic-level threats, where content may appear realistic and reasonable, yet alter the meaning of the scene. In such cases, virtual elements may adopt a consistent visual style while subtly introducing misinformation that leads to misunderstanding. Other work has focused more directly on safety concerns in AR, such as in the context of obstruction attacks\cite{attack01, attack03, obstruction01, viddar}, where virtual content blocks real-world information and potentially leads to accidents or task failures. However, while obstruction attacks involve physical concealment, they do not capture scenarios where virtual content is non-obstructive or only partially overlaps with key information, yet still alters the semantic meaning of the scene. For example, in \cref{fig:vim example 2}, additional virtual guidance is inserted into the scene, directing the user toward an incorrect action without blocking any critical real-world element.

Beyond AR-specific studies, there have been numerous studies in computer vision and multimodal machine learning that focused on scene understanding, including semantic segmentation\cite{semanticsegment02, semanticsegment01}, image captioning\cite{imagecaption01, imagecaption02}, and visual question answering (VQA)\cite{vqa01, VQA02}. These methods demonstrate impressive capabilities in extracting semantic meaning from complex visual inputs. However, they are typically developed and evaluated outside the AR context. As a result, these models' ability to analyze the presence or potential impact of augmented elements remains underexplored. In AR scenarios, this ability to analyze virtual content is important: while users can only see the augmented view, a robust detection system should evaluate both the raw scene and its augmented counterpart to determine whether virtual content has introduced detrimental changes.

These challenges call for new approaches to evaluate AR content, which go beyond visual realism or obstruction, and instead focus on the semantic impact of virtual elements. In particular, detecting whether virtual content manipulates real-world information requires understanding not just what objects are present, but how their meaning changes when combined. Recent advances in vision-language models (VLMs) with generative capabilities\cite{GPT4V, geminiteam2024geminifamilyhighlycapable, anthropic2024claude3} have demonstrated strength in multimodal reasoning, including fine-grained scene interpretation and textual analysis. Moreover, VLMs have been proven effective in evaluating artificial content\cite{VLMContentEval01} and sensitive to virtual content in AR\cite{Genaixrws}. These capabilities make VLMs well-suited for assessing whether AR content introduces semantic inconsistencies or misleading information.

Motivated by the rapid progress in VLMs and the need to detect subtle yet semantically harmful AR content, we present a comprehensive framework for identifying visual information manipulation attacks through multimodal semantic reasoning. Our approach bridges semantic reasoning with virtual content assessment, combining the strengths of VLMs and optical character recognition (OCR)-based text grounding to analyze how virtual elements alter the meaning of real-world scenes. Our key contributions are as follows:

\begin{itemize}[itemsep=0.01em]

    \item We define a taxonomy of visual information manipulation attacks in AR that categorizes manipulations by format and purpose. Specifically, we identify three manipulation formats: character, phrase, and pattern; and three attack purposes: information replacement, information obfuscation, and insertion of extra wrong information. Based on this taxonomy, we formally define seven valid types of attacks and provide a mathematical framework to describe their distinguishing characteristics.

    \item We introduce AR-VIM, a dataset of \rb{452} raw-AR video pairs collected in \rb{202} different real-world scenes. The videos are annotated to reflect various attack types and semantic distortions. To validate the dataset labels, we obtained human annotations through a user-based labeling process, approved by Duke University Campus Institutional Review Board. The results demonstrate that our labeling aligns closely with human perception. The dataset is available on GitHub.\footnote{\label{ar-vim-dataset}\url{https://github.com/YM-Xiu/AR-VIM}}

    \item We develop a multimodal semantic reasoning system, VIM-Sense, that combines VLMs with OCR-based text extraction to detect and analyze semantic-level information manipulation attacks in AR scenes. On AR-VIM, our system reaches an attack detection accuracy of \rb{88.94}\% and a detection latency of \rb{7.07} seconds. We also evaluate our system on a real-world Android AR application, achieving a detection latency of 7.17 seconds. 

\end{itemize}

The remaining sections of this paper are organized as follows: \cref{sec:related work} reviews prior work, followed by \cref{sec:VIMA}, which introduces our taxonomy of visual information manipulation attacks. \cref{sec:AR-VIM} describes the construction and annotation of the AR-VIM dataset. \cref{sec:system design} presents the design and implementation of our system, VIM-Sense. In \cref{sec:system evaluation}, we evaluate the VIM-Sense both on the AR-VIM dataset and in a real-world AR application. Finally, we discuss the limitations and directions for future work in \cref{sec:limitations} and conclude the paper in \cref{sec:conclusion}.

%% file: sections/2_related_work.tex
\section{Related Work}
\label{sec:related work}

\subsection{AR Virtual Content Assessment}


Augmented reality research has long examined how virtual content can be designed and evaluated to ensure a pleasant user experience. A significant body of work focuses on assessing the visual properties of virtual elements to determine whether they blend seamlessly into the real world. 
Kim et al.\cite{ARassessphysics01} investigated artifacts of virtual content that arise when physics simulations are used to model interactions between real and virtual objects in AR. 
Li et al. \cite{ARassessmisalign01} proposed a method for evaluating misalignment between real and virtual objects in a head-mounted AR assembly assistance system using depth map comparison. 
\rb{Golyadkin et al.\cite{inpainting01} proposed a data synthesis and benchmarking framework for manga colorization, which incorporates image inpainting techniques to support consistent colorization of sequential pages and assess the inpainting quality.}
On the user perception side, Wang et al.\cite{ARassess01} conducted a user study to investigate how texture density, luminance contrast, and color contrast of virtual objects affect depth perception in optical see-through AR. 
Liu et al.\cite{ARassesslighting02} considered visual confusion as a type of distortion affecting perceptual quality when both real and virtual contents contain complex textures. 
Duan et al.\cite{ARIQA01} proposed a framework for evaluating perceptual quality in AR scenes based on visual confusion theory.
While these works provide insights into visual realism, perceptual quality, and alignment, they primarily focus on how well virtual content fits into the physical scene or how it affects users' perception and do not analyze whether the content introduces semantic distortions or changes the meaning of the scene.


Aside from perceptual quality, some prior works have explored how virtual content can interfere with users’ access to real-world information. 
Cheng et al.\cite{attack01} conduct a user study to investigate how people perceive and react to visual obstruction attacks in AR.
Lebeck et al. proposed Arya\cite{attack03, attack02}, which detects obstructions by using system sensors to identify predefined safety-critical real-world objects. It ensures that critical items are not obstructed by modifying virtual content based on predefined policies.
In our prior work, we proposed VIDDAR\cite{viddar, viddardemo}, which integrates generative AI models with object detection models to identify critical real-world objects and determine if a key object is obstructed or made to be visually misleading.
These approaches take scene information and user safety into consideration when assessing the virtual content. However, they share a common assumption that safety threats in AR arise when important real-world objects are physically occluded. Consequently, they cannot handle more subtle cases of visual information manipulation, where virtual content does not fully block real-world objects but instead modifies or replaces their meaning.

\subsection{Scene Understanding}

Recently, numerous studies in computer vision have focused on scene understanding, which aims to extract semantic meaning from images through tasks such as semantic segmentation, image captioning, and visual question answering (VQA). Long et al.\cite{semanticsegment02} proposed fully convolutional networks for semantic segmentation, which assigns a categorical label to each pixel in an image. Vinyals et al.\cite{imagecaption01} proposed an image captioning system that uses deep neural networks to generate natural language descriptions for the image. Antol et al.\cite{vqa01} proposed the task of VQA, which requires answering open-ended questions about images using natural language. These early works laid the foundation for a wide range of scene understanding tasks. More recently, the field has seen a shift toward foundation models. Kirillov et al.\cite{SAM} proposed the Segment Anything Model (SAM), a foundation model for semantic image segmentation that supports prompt-based interaction. Generative models with visual modality, also known as VLMs, such as GPT-4V\cite{GPT4V}, Gemini\cite{geminiteam2024geminifamilyhighlycapable}, and Claude\cite{anthropic2024claude3}, have shown superior performance in some scene understanding benchmarks\cite{SUbenchmark01, vqa01, SUbenchmark02, GQA}. Overall, recent methods have demonstrated strong capabilities in understanding scene content and identifying potentially harmful or misleading information. However, most of these studies were conducted and evaluated using standard, general-purpose vision-language benchmarks. As a result, their potential for understanding virtual content in AR, where it is essential to analyze both the real-world scene and its augmented counterpart, remains largely underexplored.

\subsection{VLMs for Semantic Reasoning in AR Settings}

As mentioned above, the ability of scene understanding methods to operate effectively in AR settings remains unclear. Recently, efforts have begun to explore the use of VLMs for reasoning about artificial content such as user/AI-generated imagery, as well as virtual content presented in AR, especially in the context of safety and usability. 
Guo et al.\cite{VLMContentEval01} proposed UGCG-GUARD, a content moderation system for detecting image-based promotions of unsafe user-generated content, leveraging VLMs to perform contextual detection. 
Huang et al.\cite{hallucination} discussed the visual hallucination of VLMs on AI-generated images. 
\rb{Srinidhi et al. proposed XaiR\cite{xair}, a platform that integrates VLMs with XR headsets, which enables real-time semantic reasoning and AR content placement. Sharma et al. proposed OCTO+\cite{octo}, a framework for automatic open-vocabulary object placement in AR, which leverages VLMs to reason about spatial semantics and place virtual objects in valid regions.} Duan et al.\cite{Genaixrws} evaluated several state-of-the-art VLMs for understanding and describing AR-generated scenes, highlighting that VLMs excel at detecting and assessing obvious virtual content. Nonetheless, these work did not dive into the visual safety concerns in AR settings. In this work, we aim to fill this gap by combining VLM-based semantic reasoning with OCR-based textual grounding to detect visual information manipulation attacks in AR scenes.

%% file: sections/3_VIMA_modeling.tex
\section{Visual Information Manipulation (VIM) Attacks}
\label{sec:VIMA}

In this section, we formally define the VIM attacks in AR with mathematical models. These attacks occur when virtual content introduced into a real-world scene alters the original semantic meaning, potentially misleading users while the original content is still visible or manipulated in a not easily noticeable way. To formally describe and categorize these attacks, we consider two orthogonal dimensions: attack form, which describes how the manipulation occurs, and attack purpose, which describes what the manipulation intends to achieve. This approach follows the principles of full-reference image quality assessment (FR-IQA) methods\cite{full}, which evaluate image quality by directly comparing a distorted or modified image (in our case, the AR image $I_a$) against its original reference (the raw image $I_r$).

\subsection{Attack Format}

We categorize visual information manipulation attacks in AR based on how the virtual content alters real-world semantics.  Specifically, we define three attack formats:

\par\noindent\textbf{Character Manipulation:} Character manipulation refers to attacks where individual characters in a real-world scene are modified through virtual overlays. These changes may appear subtle (e.g., "3" to "8" or "O" to "Q"), but can significantly alter the meaning of textual information, such as room numbers, prices, or signs. 

Denote $C_r = \{c_r^1, c_r^2, \cdots, c_r^m\}$ as the ordered sequence of $m$ characters in $I_r$, and $C_a = \{c_a^1, c_a^2, \cdots, c_a^n\}$ as the ordered sequence of $n$ characters in $I_a$. These two sequences have exactly the same length, which means the characters are only manipulated, without being erased or added. Therefore, the character in $I_r$ is manipulated if:

\vspace{-0.1cm}
$$
(m=n) \land (\exists i \quad \text{s.t.} \quad c_r^i \neq c_a^i).
$$
\vspace{-0.2cm}

\par\noindent\textbf{Phrase Manipulation:} Phrase manipulation refers to attacks where a meaningful segment of text, such as a word or short phrase, is replaced or altered through virtual content in AR. Unlike character manipulation, which involves subtle changes at the letter or symbol level, phrase manipulation operates at the semantic unit level, changing the interpretation of a label, instruction, or message.

Let $P_r = \{p_r^1, p_r^2, \cdots, p_r^m\}$ be the ordered sequence of $m$ words in $I_r$ and $P_a = \{p_a^1, p_a^2, \cdots, p_a^n\}$ be the ordered sequence of $n$ words in $I_a$. Each phrase $p^i$ is a ordered sequence of characters $\{c^{i1}, c^{i2}, \cdots, c^{ij}\}$. Here, the sequences do not necessarily have the same length ($m\not\equiv n$), as phrases can be manipulated even if the new phrase has a different number of words. The phrases in $I_r$ is manipulated if:

\vspace{-0.1cm}
$$
P_r \neq P_a.
$$
\vspace{-0.3cm}

By definition, character manipulation can be viewed as a special case of phrase manipulation involving minimal changes at the character level, while also changing the phrase-level information. To avoid ambiguity, we adopt a hierarchical classification: if a manipulation satisfies both the character and phrase manipulation criteria, we categorize it as character manipulation. Only those cases that exceed the character-level threshold but still fall within the phrase-level range are classified as phrase manipulation.

\par\noindent\textbf{Pattern Manipulation:} Pattern manipulation refers to non-textual visual elements in the scene, such as colors, symbols, or directional cues, that are altered through virtual content in a way that changes the scene's semantic meaning. Unlike character or phrase manipulation, which operates on textual information, pattern manipulation affects visual structures that convey meaning independently of text.

Denote $V_r = \{v_r^1, v_r^2, \cdots, v_r^m\}$ as the ordered sequence of $m$ visual patterns in $I_r$, and $V_a = \{v_a^1, v_a^2, \cdots, v_a^n\}$ as the ordered sequence of $n$ visual patterns in $I_a$. We consider the visual patterns in $I_r$ to be manipulated if there does not exist a one-to-one mapping from $I_a$ that preserves semantic equivalence between visual elements:

\vspace{-0.2cm}
\[
\not\exists \ \phi: V_r \to V_a \quad \text{s.t.} \quad \phi(v_r^i) \overset{\text{sem}}{=} v_a^j, \ \forall i \in \{1, \dots, m\},
\]
\vspace{-0.3cm}

\noindent where $\phi$ denotes a mapping between $V_r$ and $v_a$. Note that the symbol $\overset{\text{sem}}{=}$ denotes semantic equivalence between two visual patterns. That is, two patterns are considered equivalent if they convey the same meaning, even if their pixel-level appearances differ.

\subsection{Attack Purpose}

We define the \textit{information set} of an image as the union of all meaningful semantic elements, including both textual and visual components. Formally, for a given image $I$, its information set is denoted as:
$
T = \{t_1, t_2, \dots, t_n\} = T^{\text{text}} \cup T^{\text{vis}},
$
\noindent where $T^{\text{text}}$ is the list of textual tokens present in the scene, including characters and phrases; $T^{\text{vis}}$ is the list of non-textual visual patterns. Each list is derived by mapping visual elements in the image to their corresponding semantic meanings via a human-level reasoning function $S$\cite{visualsystem01, visualsystem02}:
$
T^{\text{text}} = S(C \cap P), T^{\text{vis}} = S(V),
$
where $C$ denotes the set of all character-level elements, $P$ represents the set of all textual phrases, and $V$ denotes the set of visual patterns. The semantic reasoning function $S$ models how humans interpret these visual cues into meaningful concepts.

We denote the information sets of the raw and AR images as $T_r$ and $T_a$, respectively, where $T_r = \{t_r^1, t_r^2, \cdots, t_r^m\}$ is the ordered sequence of $m$ semantic elements in $T_r$, and $T_a = \{t_a^1, t_a^2, \cdots, t_a^n\}$ is the ordered sequence of $n$ semantic elements in $I_a$. Based on these definitions, we introduce three types of VIM attack purposes:

\par\noindent\textbf{Information Replacement:} Information replacement attack is the case where a semantic element in the raw image is removed and substituted with a different, semantically conflicting element in the AR image. An information replacement occurs if:

\vspace{-0.2cm}
\[
(m = n) \land (\exists i \quad \text{s.t.} \quad t_i^r \overset{\text{sem}}{\neq} t_i^a).
\]
\vspace{-0.2cm}

This condition indicates that the original information $t_i^r$ is no longer present in the AR image and has been replaced by another token $t_i^a$ that conveys a different semantic meaning. The substitution may occur in either the textual domain ($T^{\text{text}}$) or the visual domain ($T^{\text{vis}}$), as long as the replacement alters the perceived meaning of the original scene content.

\par\noindent\textbf{Information Obfuscation:} Information obfuscation attack refers to when a semantic element from the raw image is partially or fully occluded by virtual content in the AR image, making it difficult or impossible to interpret. Unlike information replacement, obfuscation does not introduce a conflicting semantic element; instead, it interferes with the user's ability to recognize existing information. An information obfuscation happens if:

\vspace{-0.2cm}
\[
T_a \subset T_r.
\]

This strict subset relation indicates that at least one information element present in $I_r$ is not preserved in $I_a$, and the total number of elements in $I_a$ has decreased, reflecting the loss or occlusion of visible information.

\par\noindent\textbf{Extra Wrong Information:} Extra wrong information attack is a case where new virtual content introduces semantic elements into the AR image that were not present in the raw image and that convey a misleading or incorrect meaning. Unlike information replacement or obfuscation, this type of attack does not interfere with existing real-world information but instead adds false signals that may confuse the user or alter the interpretation of the scene. An additional wrong information attack is said to occur if:

\[
(T_r \subset T_a) \land (\exists t \in T_a \setminus T_r \quad \text{s.t.} \quad \text{Contra}(t, T_r)).
\]

Here, \( T_r \subset T_a \) indicates that the AR image contains strictly more semantic elements than the raw image. Besides, at least one of the newly introduced tokens \( t \in T_a \setminus T_r \) conveys contradictory information with respect to the real-world context.

\subsection{Attack Taxonomy Summary}

\label{sec:attack summary}

Based on the definitions above, we categorize visual information manipulation attacks in AR by combining three attack formats with three attack purposes. This gives a conceptual space of $3 \times 3 = 9$ possible attack types. However, not all combinations are applicable in real-world AR scenarios. In particular, we only consider \textbf{character manipulation} in the context of information replacement, since manipulating characters at the level of obfuscation or false addition is rare or ill-defined. For example, it is difficult to obfuscate a character without affecting its surrounding word or phrase, and adding an isolated character into the scene without meaningful context rarely conveys independent semantic intent. The resulting taxonomy consists of \textbf{seven} valid attack types, as shown in ~\cref{tab:attacktaxonomy}. Each type corresponds to a unique way that virtual content can semantically alter the scene, whether through subtle text edits or high-level visual modifications. Some examples of the attacks discussed in the taxonomy are shown in \cref{fig:dataset}.


\vspace{-0.2cm}
\begin{table}[h]
\caption{Visual information manipulation attack taxonomy combining attack formats and purposes.}
\vspace{-0.2cm}
\centering
\renewcommand{\arraystretch}{1.0}
\begin{tabular}{c|ccc}
\toprule
\diagbox{Purpose}{Format} & Character  & Phrase & Pattern \\
\midrule
Replacement & \checkmark & \checkmark & \checkmark \\
Obfuscation    & --- & \checkmark & \checkmark \\
Extra Information    & --- & \checkmark & \checkmark \\
\bottomrule
\end{tabular}
\label{tab:attacktaxonomy}
\vspace{-0.4cm}
\end{table}

%% file: sections/4_AR_dataset.tex
\section{AR-VIM Dataset}
\label{sec:AR-VIM}

\begin{figure*}[t]
\includegraphics[width=1\linewidth]{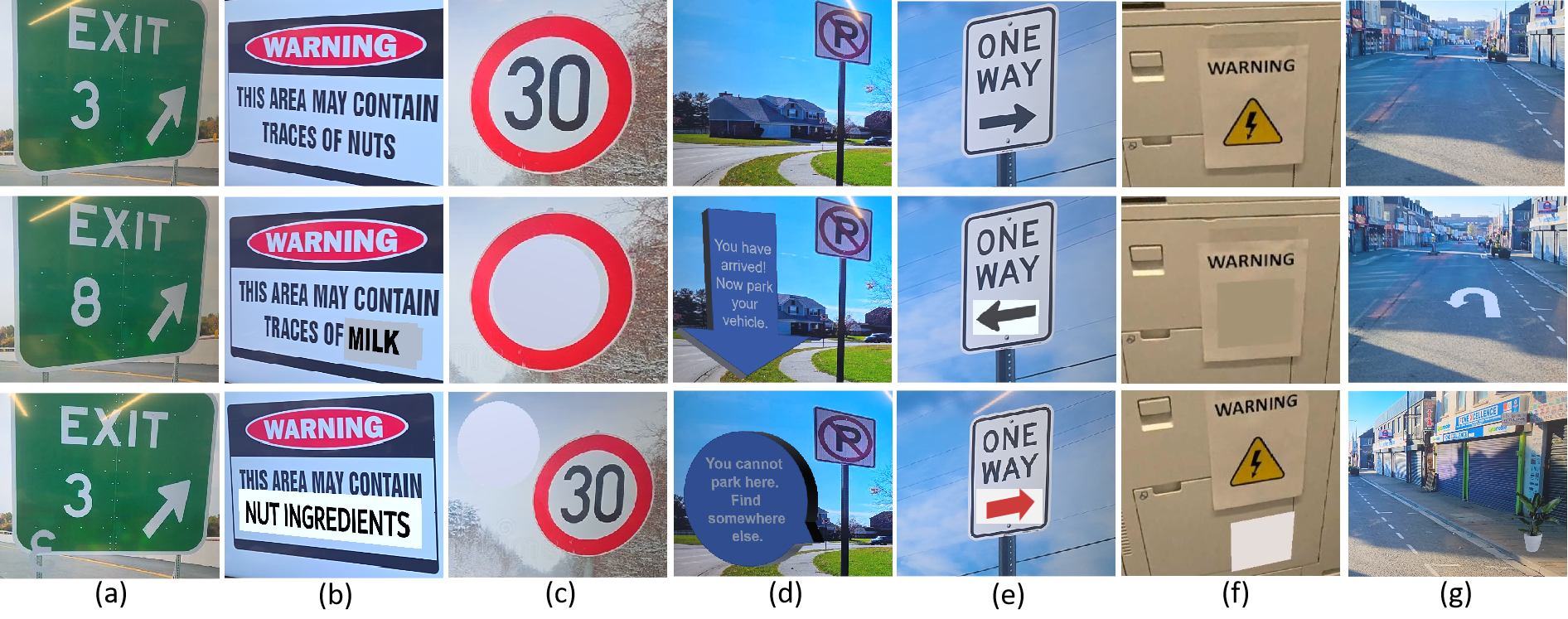}
\centering
\vspace{-0.6cm}
\caption{Samples from the AR-VIM dataset. Each column shows an attack type in a certain scene. From top to bottom: raw image, attacked AR image, and non-attacked AR image. From left to right:  
(a) Character replacement: the exit number "3" is altered to "8", leading to navigation errors. In the non-attacked case, virtual content is spatially misaligned and does not constitute an attack.
(b) Phrase replacement: "traces of nuts" is changed to "traces of milk", changing the original warning. In the non-attacked case, the phrase changes to "nut ingredients", which preserves the original warning information.  
(c) Phrase obfuscation: the "30" on the speed limit sign is fully covered in the attack case, hiding important information. In the non-attacked case, the virtual content (a white circle) is spatially misaligned, and the original sign is still visible.
(d) Phrase extra information: the attack adds misleading text encouraging the user to park near a no-parking sign. In the non-attacked case, the AR content warns the user to avoid parking in this no-parking zone.  
(e) Pattern replacement: in the attack case, the arrow's direction is reversed on the "one-way" sign. In the non-attacked case, the arrow is replaced with a highlighted one, and does not contradict the original meaning.
\rb{(f) Pattern obfuscation: an electricity warning symbol is hidden in the attack case, removing a critical safety indicator. In the non-attacked case, the virtual content is spatially misaligned while the symbol remains fully visible. }
(g) Pattern extra information: the attack introduces a U-turn road sign to the scene, misleading users that they can only U-turn here. In the non-attacked case, a clearly contextually irrelevant and semantically neutral object (a plant) is added.}

\label{fig:dataset}
\vspace{-0.2cm}
\end{figure*}

\rb{
\subsection{VIM Attack Design Methodology}

 Based on the VIM attack taxonomy, we aim to build a comprehensive dataset of VIM attacks in AR as a standard benchmark. To simulate realistic VIM attacks, we adopted a systematic design methodology. Our goal was to construct attack cases that are subtle yet semantically impactful while maintaining a high degree of visual coherence with the surrounding environment. Each attack instance was manually crafted through a scene-specific design process guided by the following principles:

\noindent \textbf{Semantic Alteration with Minimal Virtual Content Insertion}: 
For each scene, our objective was to introduce the minimal amount of virtual content required to alter the scene’s meaning. For example, in character replacement, we selected individual characters whose modification would flip the semantic interpretation (e.g., "F" to "E" and "1" to "7") while leaving the surrounding context unchanged. This ensures the manipulation is subtle yet capable of misleading the user.

\noindent \textbf{Visual Style Coherence}:
To avoid triggering user attention, we ensured that the visual style of the virtual content closely matched the real-world context, which includes color style, lighting effect, and spatial alignment. This is especially important for information replacement and information obfuscation attacks, where the effectiveness of the manipulation relies on the user's inability to recognize the virtual content. For extra wrong information attacks, we focused on adding semantically misleading but visually plausible content.

\noindent \textbf{Attack-Non-Attack Pairing Strategy}:
To ensure controlled comparisons and reliable evaluation of VIM detection, we designed the corresponding non-attacked case for most of the attacked cases. For information obfuscation, we created a non-attacked version by repositioning the virtual content to a different part of the scene that does not obfuscate the key semantic elements in the real world. For information replacement and extra wrong information attacks, we generated visually similar virtual content in terms of color, style, and size, but with either misaligned placement or semantically innocuous meaning. For example, in ~\cref{fig:dataset}(b), the attack changes "traces of nuts" to "traces of milk", and the non-attacked version might replace "traces of nuts" with "nut ingredients" in the same font and location, preserving visual realism while avoiding deception.
}



\subsection{Data Collection Pipeline}

\begin{table*}[t]
\caption{\rb{Data distribution of AR-VIM dataset by attack types and attack labels.}}
\vspace{-0.2cm}
\centering
{
\renewcommand{\arraystretch}{1}
\begin{tabular}{c|c|c|c|c|c|c|c|c}
\toprule
\textbf{Attack Type} &
\makecell{Character \\ Replacement} &
\makecell{Phrase \\ Replacement} &
\makecell{Phrase \\ Obfuscation} &
\makecell{Phrase Extra \\ Information} &
\makecell{Pattern \\ Replacement} &
\makecell{Pattern \\ Obfuscation} &
\makecell{Pattern Extra \\ Information} &
\makecell{Total}\\
\midrule
\textbf{Attacked}     & 32 & 34 & 31 & 40 & 39 & 31 & 34 & 241\\
\textbf{Non-Attacked}   & 32 & 28 & 27 & 40 & 28 & 28 & 28 & 211\\
\midrule
\textbf{Total}        & 64 & 62 & 58 & 80 & 67 & 59 & 62 & 452\\
\bottomrule
\end{tabular}
}
\label{tab:dataset_distribution}
\vspace{-0.4cm}
\end{table*}


\rb{With the attack design methodology in place, we designed two controlled data collection pipelines that simulate a wide range of real-world scenarios. One pipeline places AR content onto static images displayed on a monitor, capturing the scene through a smartphone to collect AR data. This introduces a more diverse range of scenarios and reduces data collection efforts by eliminating the need for data collectors to be physically present in the locations described in the background images. The other pipeline deploys the AR content directly in real-world environments using an AR headset, capturing video from the user's view that reflects practical usage conditions. This setup enables us to balance scenario diversity with realism.}

\rb{\noindent\textbf{Monitor-Based AR Data Collection}:} We started by collecting 58 high-resolution real-world scene images from online sources and generative AI models. \rb{The AI-generated backgrounds are a double-edged sword: they enable a broader range of scenarios that are often difficult to capture in the real world, but may also introduce domain shifts that may affect the behavior of VLMs.} All of these backgrounds form a diverse set of environments such as street signs, building facades, hospital entrances, and parking instructions. We display these images on a monitor one by one as the background. For each background, we manually designed virtual content to simulate either attacked or non-attacked cases. The content varies across attack formats and purposes, based on the taxonomy introduced in \cref{sec:VIMA}. These assets were imported into Unity and positioned at designated regions. Then, we implemented an AR application for Android phones using Unity with ARCore support. Specifically, we used ARCore's image tracking capability to precisely align the virtual content with the background scene. In our setup, each background image is used as a reference image. When the phone’s camera detects the reference image at runtime, ARCore registers its spatial pose and enables Unity to attach virtual content to the same coordinate system. After that, we recorded each AR experience using two virtual AR cameras in Unity. One camera renders only the background scene to produce the raw image \( I_r \), while the other renders both the background and the virtual content to produce the AR image \( I_a \). Finally, we combined sequences of raw/AR images into raw/AR video pairs, each pair representing a specific scenario.

We used a Samsung Galaxy S25 smartphone to record the videos and a 55-inch 4K Samsung monitor to display the background scenes. The AR app was developed in Unity 2022.3.28f1 with ARCore for image tracking and rendering.

\rb{
\noindent\textbf{Real-World AR Data Collection}: While the monitor-based pipeline allows for efficient prototyping and introduces a more diverse range of scenarios, it may fail to capture important spatial properties of real-world AR environments such as 3D depth cues, which are intrinsic to real AR environments. To address this limitation, we additionally collected some data in real-world settings. Specifically, we developed an AR application for the Meta Quest 3 in Unity 2022.3.61f1, which supports access to the device’s main camera. The app allows users to grab and place virtual content at desired locations using the controller. We manually designed 35 combinations of real-world scenes and virtual content to comprehensively cover all attack formats and purposes described in \cref{sec:VIMA}. The main camera of the Meta Quest 3 was accessed to capture raw images \( I_r \), while a virtual camera in Unity rendered the virtual content separately. The virtual content was then overlaid onto the raw images to produce the corresponding AR images \( I_a \). Finally, we also combined image sequences to create the videos.

}

\subsection{Dataset Composition}

With the data collection pipelines, we developed the AR-VIM dataset. The final AR-VIM dataset consists of \rb{452} video pairs, each containing a raw video and its corresponding augmented version. These pairs span a total of \rb{202} unique scenes of AR experiences. \rb{Specifically, 307 video pairs with 133 scenes were collected with the monitor-based pipeline, while 145 video pairs with 69 scenes were collected using the real-world pipeline.} Each pair simulates either an attacked or non-attacked scenario using the taxonomy introduced in \cref{sec:VIMA}. For each video pair, the virtual content appears at a specific timestamp, simulating a real-time AR interaction where the virtual overlay is introduced during scene observation. Each video in the dataset has a resolution of \rb{either 480×1080 pixels (monitor-based videos) or 960×1280 pixels (real-world videos)}, and a duration ranging from 4 to 17 seconds. The videos had a framerate of 15 FPS, a constraint imposed by smartphone hardware limitations, as running an AR application while simultaneously capturing high-frame-rate raw and AR videos places strain on AR devices.

To ensure balanced coverage across both manipulation formats and purposes, we constructed seven types of attacks, as elaborated upon in ~\cref{sec:attack summary}: character replacement, phrase replacement, phrase obfuscation, phrase extra information, pattern replacement, pattern obfuscation, and pattern extra information. For each attack type, we also generated a comparable number of non-attacked cases for comparison. In total, the dataset includes \rb{241 attacked video pairs and 211 non-attacked video pairs}, ensuring a balanced distribution, as shown in ~\cref{tab:dataset_distribution}.

Some of the data examples, together with the attack labels, are shown in \cref{fig:dataset}. The full dataset, including video pairs, original image frames, and attack labels, is publicly released on Github\footref{ar-vim-dataset}.

\subsection{User-Based Label Validation}

To ensure that the attack labels in our dataset align with human perception, we conducted an IRB-approved user-based labeling task to validate our annotations. An interactive web interface was developed using Gradio~\cite{gradio}, where each participant was shown a randomly selected raw-AR video pair from the AR-VIM dataset. Prior to starting the task, participants received a brief overview of VIM attacks in AR to calibrate their understanding. After viewing each video pair, they were asked to indicate their level of agreement with the statement: "\textit{The AR video contains a VIM attack}" on a 5-point Likert scale~\cite{likert}, ranging from 1 (Strongly Disagree) to 5 (Strongly Agree). For non-attacked cases, where disagreement indicates correct perception, we inverted the scores by subtracting them from 6, ensuring that higher values consistently indicate stronger agreement with the correct label. The user interface for data validation is shown in ~\cref{fig:ui}.

We recruited \rb{26} participants \rb{(aged 21 to 56; 8 female)}, with each participant evaluating 40 video pairs that were randomly selected to ensure a balanced distribution of attack labels and types. To support both in-person and remote participation, the data labeling task was conducted in a hybrid format: in-person participants viewed videos on a 16-inch laptop, while remote participants used the same interface displayed on their personal devices. 

The distribution of agreement scores is shown in ~\cref{fig:chart}. \rb{The average agreement score across the dataset is 4.53}, indicating strong alignment between the dataset labels and human judgment. We also observed that visual pattern-related videos received relatively lower agreement scores, likely because humans have a more subjective and diverse understanding of visual patterns compared to text. The user-based data labeling results validate the perceptual soundness of AR-VIM and confirms that VIM attacks, as defined in our taxonomy, are generally identifiable by human users.

%% file: sections/5_system_design.tex
\section{System Design}
\label{sec:system design}

\begin{figure}[t]
\includegraphics[width=1.0\linewidth]{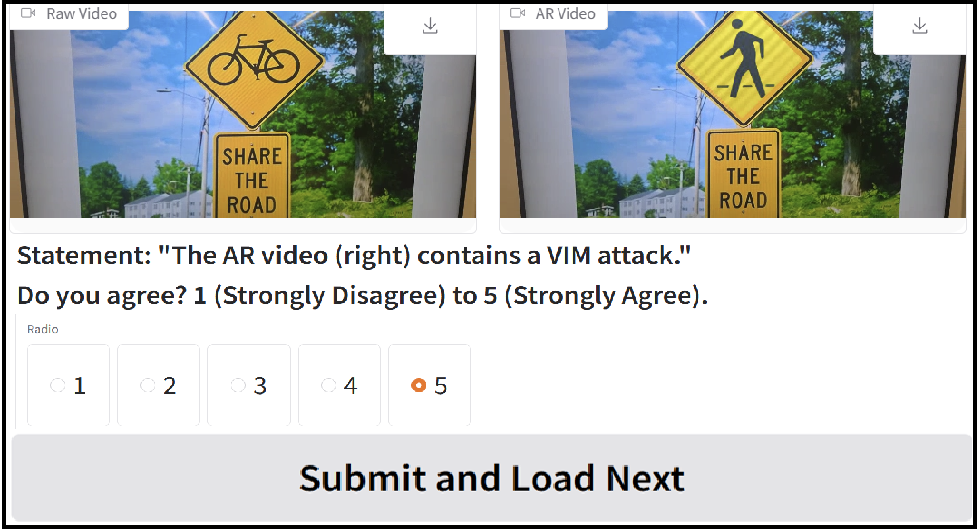}
\centering
\vspace{-0.2cm}
\caption{Interface used in the user-based labeling task for evaluating visual information manipulation in AR videos.}
\label{fig:ui}
\vspace{-0.2cm}
\end{figure}

\begin{figure*}[t]
\includegraphics[width=1.0\linewidth]{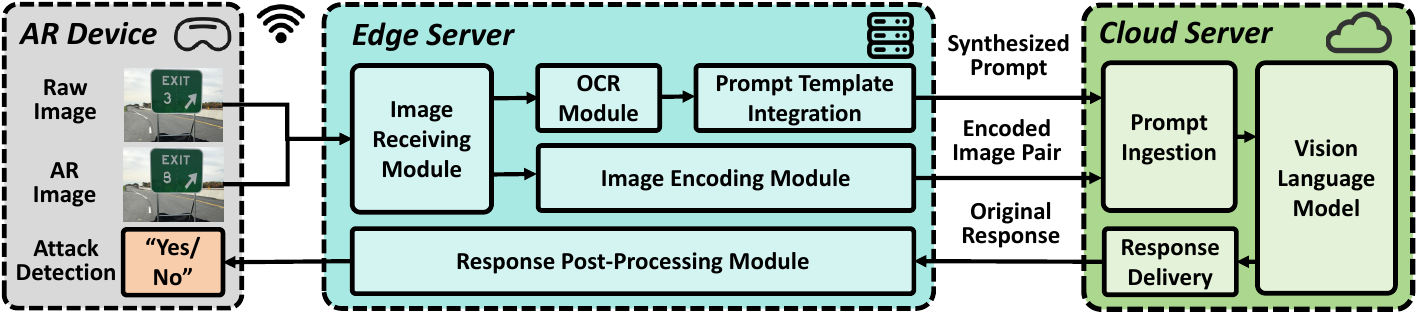}
\centering
\vspace{-0.2cm}
\caption{System architecture of VIM-Sense.}
\label{fig:vim diag}
\vspace{-0.55cm}
\end{figure*}

To detect VIM attacks in AR, we propose VIM-Sense, a multimodal reasoning system that operates across an edge-cloud architecture. As shown in \cref{fig:vim diag}, the system takes a raw and AR image pair as input, and outputs a binary prediction indicating whether an attack has occurred. VIM-Sense consists of two major components: (1) an edge server responsible for lightweight pre-processing, prompt construction, and post-processing; (2) a cloud server that performs semantic reasoning using a VLM. The edge server accesses the models on the cloud over the internet via Wi-Fi.

\noindent \textbf{Edge Server:}
The edge server acts as the first stage of processing and handles lightweight yet essential tasks before invoking the more computationally intensive reasoning of the cloud-based VLM. This module is deployed on local devices or nearby servers to minimize delay. Once the system receives a new input, the image receiving module ensures the raw and AR images are successfully captured, temporally synchronized, and spatially aligned. Following image acquisition, the OCR module is employed to extract visible textual elements from both images. This includes characters, phrases, or other meaningful symbols. The OCR output is combined with a prompt template module. This prompt is designed to guide the downstream VLM through a series of reasoning steps to determine whether a manipulation has occurred. A sample prompt is shown below:

\begin{tcolorbox}[colback=green!10!white, 
                  colframe=black, 
                  boxsep=4pt,  
                  left=4pt,    
                  right=4pt,   
                  top=4pt,     
                  bottom=4pt   
                 ]
\noindent "You are an expert in evaluating augmented reality scenes. I will show you two images: Image 1: a photo of the real-world scene; Image 2: an augmented version of the same scene with virtual content overlaid.

\noindent \textbf{The text in the real-world scene is \{raw\_text\} and the augmented scene is \{ar\_text\}.}

\noindent Your task is to analyze how the virtual content in Image 2 affects the perception of the original scene. Please answer the following questions one by one:

\noindent 1. What is the real-world scenario shown in Image 1? Provide a brief description of the environment and the objects present.

\noindent 2. Describe the virtual content added in Image 2. What are the virtual objects, where are they placed, and how do they visually interact with the original scene?

\noindent 3. Does the virtual content alter or interfere with the original objects or information in the real-world scene (Image 1)? If so, how might this affect someone interpreting or interacting with the scene? You can assume the user knows this is an AR scene, but does not know the virtual objects in advance, so he will NOT be affected by virtual content that is obvious or not seamlessly blended into the scene. If the information in the scene is changed to something similar to its original meaning, it is fine and should not be considered manipulated. The user is not so easy to be fooled or get confused.

 \noindent 4. Based on your analysis in Question 3, can this virtual content potentially lead to misunderstanding, distraction, or harm? If yes, answer only with `Yes'. Otherwise answer `No'."
\end{tcolorbox}

The prompt is structured to encourage step-by-step, chain-of-thought reasoning\cite{cot}, guiding the VLM through a multi-stage analysis of the scene. Rather than directly asking for a binary classification, the model is prompted to first understand the real-world context, identify and describe the virtual content, assess how it affects the original information, and finally make a judgment about the attack. The prompt also explicitly considers user awareness and cognitive resilience, instructing the model to disregard virtual elements that are visually obvious or stylistically inconsistent with the scene, thereby reducing false positives. The bold portion of the prompt is dynamically generated based on the OCR results. If no text is detected in either $I_r$ or $I_a$, the bold segment is replaced with:

\begin{tcolorbox}[colback=green!10!white, 
                  colframe=black, 
                  boxsep=4pt,  
                  left=6pt,    
                  right=4pt,   
                  top=4pt,     
                  bottom=4pt   
                 ]
"\textbf{In these images there is no text.}"
\end{tcolorbox}

\noindent And if the text in both images is consistent, a general semantic reasoning prompt is used to analyze visual patterns alone, by replacing the bold text in the prompt with:

\begin{tcolorbox}[colback=green!10!white, 
                  colframe=black, 
                  boxsep=4pt,  
                  left=6pt,    
                  right=4pt,   
                  top=4pt,     
                  bottom=4pt   
                 ]
    "\textbf{The text has not changed between the two images.}"
\end{tcolorbox}

After the prompt is constructed, the image encoding module encodes the image pair into Base64 format. Alongside the prompt, the encoded image pair is sent to the cloud server via an API. After the cloud server processes the task and returns the generated response, the edge server’s response post-processing module extracts the final prediction from the rich natural language explanations.  We search the model’s output for the last occurrence of "Yes" or "No," use the result as the final attack prediction, and return it to the AR interface.

\begin{figure}[t]
\includegraphics[width=1.0\linewidth]{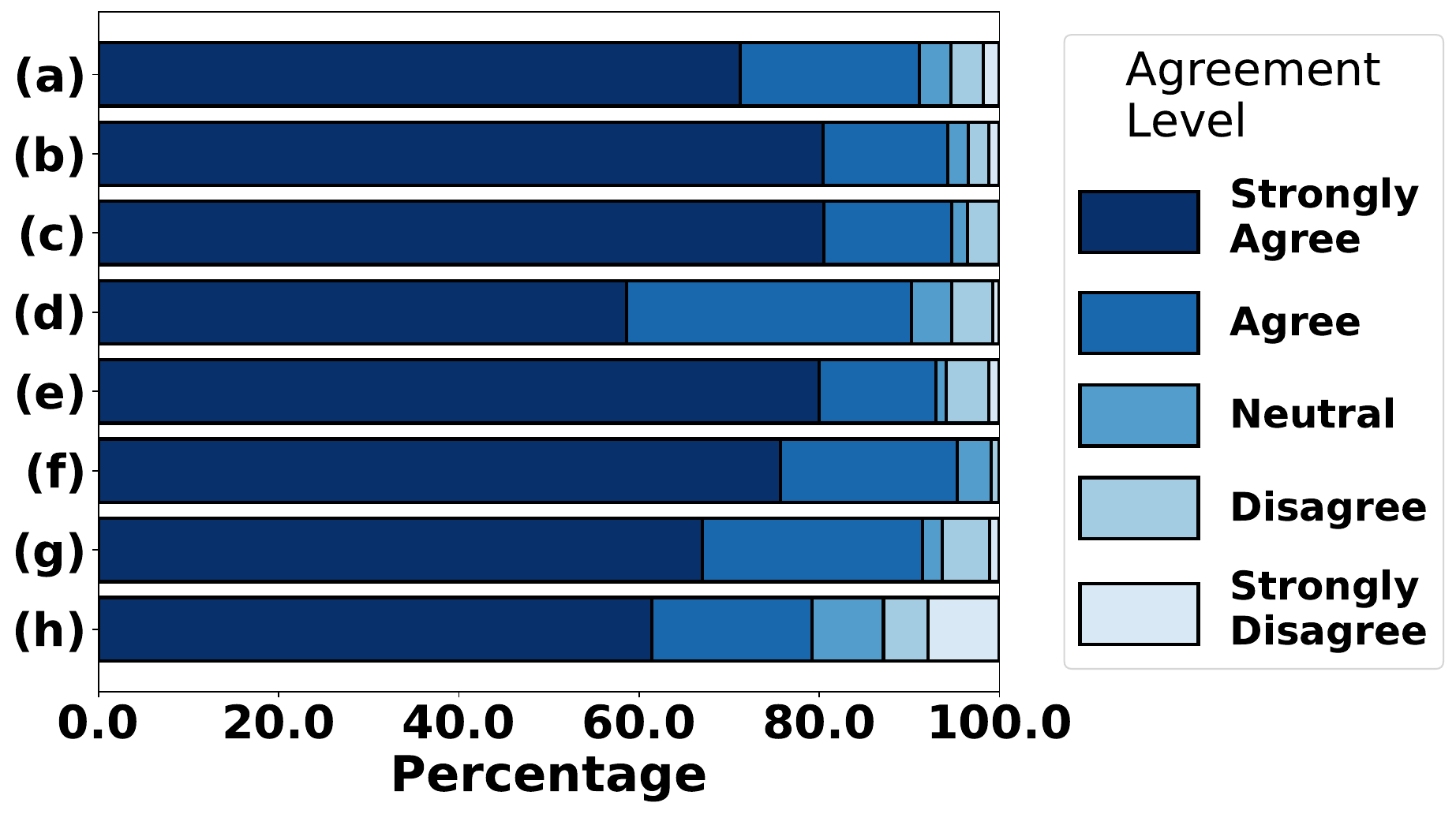}
\centering
\vspace{-0.3cm}
\caption{\rb{User agreement with attack labels in the AR-VIM dataset.} (a) The overall distribution of Likert-scale responses. (b)-(h) Likert responses for all seven attack types:
(b) Character replacement,
(c) Phrase replacement,
(d) Phrase obfuscation,
(e) Phrase extra information,
(f) Pattern replacement,
(g) Pattern obfuscation,
(h) Pattern extra information.
}

\label{fig:chart}
\vspace{-0.6cm}
\end{figure}

\noindent \textbf{Cloud Server:} The cloud server hosts a VLM, which serves as the core semantic reasoning engine in VIM-Sense. Upon receiving the structured prompt and encoded image pair from the edge server, the VLM jointly processes the textual and visual information to complete the multi-step reasoning task. The model’s response includes a detailed analysis of the scene and concludes with a binary decision, which is subsequently extracted and interpreted by the edge server to finalize the attack prediction.

%% file: sections/6_evaluation.tex
\section{System Evaluation}
\label{sec:system evaluation}

\subsection{Experiment Setup}

To evaluate the effectiveness of VIM-Sense, we conducted a series of experiments on the AR-VIM dataset. The evaluation is designed to assess the system’s ability to detect VIM attacks across diverse scenarios and compare its performance against multiple baselines.




\noindent \textbf{Proposed Method:}
Our system, VIM-Sense, combines VLMs with OCR-based prompt construction and post-processing to identify attacks in raw-AR image pairs. For all experiments, we use the full pipeline described in \cref{sec:system design} unless otherwise noted. We test VIM-Sense using two state-of-the-art commercial VLMs, GPT-4o by OpenAI and Gemini-1.5-Pro by Google, \rb{as well as an open-source VLM, LLaVA-OneVision\cite{llavaonevision}.} For the OCR component, we employed the EasyOCR library implemented in Python, a deep learning-based OCR engine that supports text recognition in multiple languages.

\noindent \textbf{Baselines:}
To contextualize the performance of VIM-Sense, we compare it with the following baselines:

\begin{itemize}[itemsep=0.01em]

    \item \textbf{GenAI-only}: A simplified version of our system that skips the OCR module and uses only a generic prompt to let the VLM analyze visual content without textual cues.

    \item \textbf{GenAI-Underdetailed}: A further reduced version where the prompt omits detailed guidance and only asks if there is any visual information manipulation in the AR scene, without staged reasoning or chain of thought.

    \item \textbf{OCR-only}: A rule-based baseline that detects attacks by comparing the textual content extracted from the raw image $I_r$ and the AR image $I_a$ using the EasyOCR module. This approach does not incorporate semantic reasoning or any visual context beyond textual content. We define a text preservation threshold of 0.9: if fewer than 90\% of the textual elements from $I_r$ are preserved in $I_a$, the system flags the pair as an attack. For fairness, we do not evaluate this baseline on pattern manipulation tasks, as these do not involve textual changes.

    \item \textbf{Feature Similarity}: A visual feature-based baseline that encodes both $I_r$ and $I_a$ using a pre-trained vision encoder and computes the cosine similarity between their global embeddings. A threshold of 0.9 is applied to classify semantic changes: if the similarity between the embeddings is lower than the threshold, it may indicate the AR image is significantly different from the raw image, and a VIM attack is likely to occur.

\end{itemize}


In the experiments, we use GPT-4o-2024-08-06, Gemini-1.5-Pro-002, and \rb{LLaVA-OneVision-7B} as the VLMs. GPT-4o and Gemini-1.5-Pro are accessed via commercial APIs provided by OpenAI and Google, respectively, \rb{while LLaVA-OneVision is deployed on Google Colab.} For other modules, we adopt EasyOCR 1.7.2 as the OCR module and CLIP 1.0 with a ViT-B/32 backbone \cite{vit} as the vision encoder.

\subsection{Results}

\begin{table*}[ht]
\caption{\rb{VIM attack detection accuracy and latency of VIM-Sense and baselines.}}
\vspace{-0.2cm}
\renewcommand{\arraystretch}{1.1}
\setlength{\tabcolsep}{5pt}
\centering
\begin{tabular}{c|c|ccccccc|c|c}
\toprule
\multirow{3}{*}{\makecell{Detection \\ Method}} & 
\multirow{3}{*}{\makecell{VLM}} & \multicolumn{8}{c|}{Detection Accuracy (\%)} & \multirow{3}{*}{\makecell{Detection \\ Latency \\ (second)}} \\
\cmidrule(lr){3-10}
& & 
\makecell{Character \\ Repl.} & 
\makecell{Phrase \\ Repl.} & 
\makecell{Phrase \\ Obfus.} & 
\makecell{Phrase \\ Extra Info.} & 
\makecell{Pattern \\ Repl.} & 
\makecell{Pattern \\ Obfus.} & 
\makecell{Pattern \\ Extra Info.} & 
\makecell{Overall}\\
\midrule
\multirow{3}{*}{\textbf{VIM-Sense}} 
& GPT-4o         & \textbf{85.94} & \textbf{95.16} & 86.21 & \textbf{92.50} & \textbf{92.54} & \textbf{91.53} & \textbf{77.42} & \textbf{88.94} & 7.07 \\
& Gemini-1.5-Pro  & 76.56 & 88.71 & 63.79 & 75.00 & 89.55 & 67.80 & 61.29 & 75.00 & 5.57 \\
& LLaVA-OV-7B  & 51.56 & 61.29 & 48.28 & 60.00 & 49.25 & 54.24 & 56.45 & 54.65 & 9.49 \\
\midrule
\multirow{3}{*}{GenAI-Only} 
& GPT-4o         & 84.38 & 90.32 & 82.76 & 91.25 & 89.55 & \textbf{91.53} & \textbf{77.42} & 86.95 & 6.96 \\
& Gemini-1.5-Pro  & 73.44 & 83.87 & 60.34 & 72.50 & 86.57 & 62.71 & 59.68 & 71.68 & 5.42 \\
& LLaVA-OV-7B  & 45.31 & 48.39 & 53.45 & 52.50 & 46.27 & 50.85 & 41.94 & 48.45 & 8.92 \\
\midrule
\multirow{3}{*}{\makecell{GenAI-\\Underdetailed}}
& GPT-4o         & 56.25 & 72.58 & 50.00 & 76.25 & 59.70 & 61.02 & 59.68 & 62.83 & 3.35 \\
& Gemini-1.5-Pro  & 53.13 & 59.68 & 53.45 & 76.25 & 59.70 & 54.24 & 53.23 & 59.29 & 1.96 \\
& LLaVA-OV-7B  & 50.00 & 45.16 & 48.28 & 55.00 & 38.81 & 45.76 & 50.00 & 47.79 & 1.26 \\
\midrule
OCR-Only                 & N/A & 71.88 & 48.39 & \textbf{89.66} & 51.25 & N/A & N/A & N/A & 64.02 & 1.05 \\
\midrule
Feature Similarity       & N/A & 50.00 & 46.77 & 62.07 & 55.00 & 47.76 & 66.10 & 46.77 & 53.32 & \textbf{0.81} \\
\bottomrule
\end{tabular}
\vspace{-0.4cm}
\label{table:main results}
\end{table*}

We evaluate system performance using two key metrics: attack detection accuracy and latency.

\noindent \textbf{Accuracy}: Since VIM-Sense performs binary classification (attack or non-attack), we adopt classification accuracy as our primary metric. It is defined as the percentage of video pairs correctly identified with respect to their ground truth labels. This metric directly reflects a method's capability in detecting VIM attacks in AR.

\noindent \textbf{Latency}: To better reflect real-world usage where AR scenes evolve over time, we simulate dynamic image transmission flows using video input. Specifically, each video pair is sampled every 0.5 seconds to get the raw frame and AR frame. When a difference is detected between the sample raw and AR image frames, it indicates that virtual content has appeared, triggering the attack detection pipeline. We then measure the time between the timestamp where the virtual content appears and the moment the system returns its binary prediction. This duration is reported as the detection latency, which captures the system’s responsiveness in practical AR settings.

~\cref{table:main results} summarizes the detection accuracy of VIM-Sense compared with several baselines across all seven attack types, as well as the average detection latency. VIM-Sense achieves the best detection accuracy across all seven attack types, reaching an overall accuracy of \rb{\textbf{88.94\%} when using GPT-4o. It performs particularly well on phrase replacement (95.16\%). Even in relatively low-performing categories such as pattern extra information, the system maintains a competitive performance (77.42\%). These results highlight the effectiveness of combining textual grounding with semantic reasoning. Using Gemini-1.5-Pro, the overall accuracy is lower (75.00\%), but still competitive given the model's different reasoning capabilities and sensitivity to virtual content in AR. When using LLaVA-OneVision, the accuracy dropped significantly to 54.65\%, suggesting that moderate-scale, resource-friendly open-source VLMs are currently insufficient for effectively detecting VIM attacks.}

In terms of latency, VIM-Sense requires \rb{\textbf{7.07 seconds} with GPT-4o, \textbf{5.57 seconds} with Gemini-1.5-Pro, and \textbf{9.49 seconds} with LLaVA-OneVision} to return a decision after detecting virtual content. This reflects the additional processing time required for image encoding, OCR extraction, and especially multi-stage prompting. Nevertheless, it remains within an acceptable range for many AR applications, particularly in scenarios where the environment is relatively static and does not require real-time responsiveness.

We also analyze the performance of baselines in detail:

\begin{itemize}[itemsep=0.01em]
    \item \textbf{GenAI-Only}: This baseline removes the OCR-based text grounding, relying solely on generic prompts for semantic reasoning. While it also achieves a high accuracy of \rb{86.95\%} (GPT-4o), which is less than 2\% lower than our proposed method, its performance on certain attack types, especially those related to text modality like phrase replacement and phrase obfuscation degrades slightly, showing that grounding the VLM in scene text improves sensitivity to text and results in a better system performance. Latency is reduced to \rb{6.96 seconds} due to simplified pipeline construction.

    \item \textbf{GenAI-Underdetailed}: This baseline simplifies the prompt further by completely removing reasoning steps and directly asking the VLM if the AR content is manipulative. The lack of instruction clarity leads to a significant drop in performance (\rb{62.83}\% for GPT-4o, more than \rb{26}\% lower than the proposed method.) This confirms the importance of chain-of-thought guidance in understanding subtle semantic changes. It also yields a much lower latency \rb{(3.35 seconds with GPT-4o, 1.96 seconds with Gemini and 1.26 seconds with LLaVA-OneVision)} because the reasoning steps are largely reduced.


    \item \textbf{OCR-Only}: This baseline relies on a simple text preservation threshold to flag potential attacks. While it does not incorporate any semantic reasoning or visual pattern analysis, it surprisingly achieves a moderate overall accuracy of \rb{64.02}\%, which outperforms all the GenAI-Underdetailed baselines. \rb{Notably, it outperforms VIM-Sense in phrase obfuscation tasks. This is because in phrase obfuscation cases, the text in the images is usually largely obstructed, while in the non-obfuscated cases, the text remains largely unchanged, making this task especially well-suited for the OCR-based approach.} However, it still underperforms in all other tasks compared to the proposed method, as those tasks typically involve more semantic changes without significantly altering the length of text. With a latency of \rb{1.05 seconds}, OCR-Only is very fast due to its simple logic and not employing a VLM.

    \item \textbf{Feature Similarity}: Here we encode global features from both images and compare them using cosine similarity. Despite being vision-based, this baseline lacks semantic awareness and performs weakly across all attack types, resulting in the lowest accuracy (\rb{53.32}\%). It has the lowest latency (0.81 seconds) since it only involves feature extraction and vector comparison.
\end{itemize}



\subsection{Ablation Study}

To better understand the role of prompt design in the performance of VIM-Sense, we conduct an ablation study focusing on the system's sensitivity to potential VIM attacks. Specifically, we examine whether adjusting the language instructions used in the prompt to reflect different assumptions about user vulnerability can influence the model’s predictions.

In the default prompt described in \cref{sec:system design}, we instruct the VLM to reason under the assumption that the user is moderately resilient to deception. That is, the user is aware they are in an AR scene and will not be misled by virtual content that is stylistically inconsistent or has minimal semantic deviation from the real scene. This is reflected in the following part of the prompt template:

\vspace{-0.1cm}
\begin{tcolorbox}[colback=green!10!white, 
                  colframe=black, 
                  boxsep=4pt,  
                  left=4pt,    
                  right=4pt,   
                  top=4pt,     
                  bottom=4pt   
                 ]
    "You can assume... it is fine and should not be considered manipulated. \textbf{ The user is not so easy to be fooled or get confused.}"
\end{tcolorbox}
\vspace{-0.1cm}

To explore the impact of this design choice, we evaluate two ablated versions of the prompt for attack detection. In ablation 1, we remove the sentence:

\vspace{-0.1cm}
\begin{tcolorbox}[colback=green!10!white, 
                  colframe=black, 
                  boxsep=4pt,  
                  left=4pt,    
                  right=4pt,   
                  top=4pt,     
                  bottom=4pt   
                 ]
    "\textbf{The user is not so easy to be fooled or get confused.}"
\end{tcolorbox}
\vspace{-0.1cm}

\noindent from the prompt. This eliminates the guidance about the cognitive strength of the user, making the model more neutral. In ablation 2, we replace the entire cognitive resilience statement with: 

\begin{tcolorbox}[colback=green!10!white, 
                  colframe=black, 
                  boxsep=4pt,  
                  left=4pt,    
                  right=4pt,   
                  top=4pt,     
                  bottom=4pt   
                 ]
    "You can assume the user knows this is an AR scene, but does not know the virtual objects in advance, 
    \textbf{so the user is easy to be fooled or gets confused.}"
\end{tcolorbox}

\noindent This version assumes a highly vulnerable user, encouraging VIM-Sense to treat even minor changes as potential VIM attacks. 

We evaluate these prompt versions on the AR-VIM dataset using the GPT-4o model and the full GenAI + OCR pipeline. As shown in ~\cref{table:ablation}, we report four classification metrics: true positives (TP, correctly predicted attacks), true negatives (TN, correctly predicted non-attacks), false positives (FP, non-attacks misclassified as attacks), and false negatives (FN, attacks missed by the system).

The standard prompt achieves the best overall performance, with an accuracy of \rb{88.94\%, 215 TP, and 187 TN.} In ablation 1, removing the cognitive resilience statement leads to a slight increase in FN and a significant increase in FP, resulting in a \rb{4.4}\% drop in accuracy. This indicates that a minor relaxation of user modeling can reduce the system’s performance. In ablation 2, where the prompt instructs the VLM to assume that users are easily confused, the model becomes overly sensitive to visual changes. While TP increases slightly to \rb{228, TN drops significantly to 86, and FP jumps to 125}, indicating that the system is misidentifying many non-attacked cases as attacks. This shift leads to the worst accuracy (\rb{69.47}\%). These findings highlight the importance of careful prompt design in affecting the system’s sensitivity to potential attacks. Overly cautious prompts can trigger false alarms, while under-specified prompts may fail to capture subtle yet critical manipulations.


\vspace{-0.0cm}
\begin{table}[h]
\caption{\rb{VIM attack detection results for different prompts.}}
\vspace{-0.2cm}
\renewcommand{\arraystretch}{1.1}
\centering
\begin{tabular}{c|cccc|c}
\toprule
\textbf{Prompt Type} & \textbf{TP} & \textbf{TN} & \textbf{FP} & \textbf{FN} & \textbf{Accuracy (\%)}\\
\midrule
Standard & 215 & 187 & 24 & 26 & 88.94 \\
\midrule
Ablation 1 & 212 & 170 & 41 & 29 & 84.51 \\
\midrule
Ablation 2 & 228 & 86 & 125 & 13 & 69.47 \\
\bottomrule
\end{tabular}
\label{table:ablation}
\vspace{-0.5cm}
\end{table}

\subsection{Real-World Deployment and Latency Measurement}

To evaluate the practicality and responsiveness of VIM-Sense in a real-world setting, we modify the Android AR application developed for \rb{monitor-based} dataset collection. Specifically, we introduce a simple user interface that allows users to manually trigger attack detection. During testing, users are presented with AR scenarios rendered live on the phone. Upon observing the scene, they press a detection button to initiate the VIM-Sense pipeline. Once triggered, the system captures the raw and AR images, processes them on the edge server, and sends the synthesized prompt and image encodings to the cloud-based VLM for reasoning. The response is then returned to the edge device. After post-processing, the binary detection result is displayed on the screen. We measure the system's end-to-end latency as the time elapsed between the user clicking the detection button and the display of the final attack prediction. A screenshot of our real-world AR app implementation is shown in ~\cref{fig:real world ui}.

The real-world testing is conducted using the same Samsung Galaxy S25 smartphone for AR rendering and a local workstation acting as the edge server, equipped with an NVIDIA RTX 3090 GPU. For cloud-based reasoning, we continue to use OpenAI’s GPT-4o as the backend VLM. A total of 30 trials were performed in an indoor environment where the smartphone and edge server communicated over a one-hop Wi-Fi connection. The average detection latency measured is \textbf{7.17 seconds}, slightly higher than the \rb{7.07} seconds observed in our simulated detection setup, largely due to the data transmission overhead between the AR device and the edge server in real-world conditions. Nonetheless, the results confirm that VIM-Sense is capable of functioning interactively in real-time AR environments, delivering timely feedback on potential VIM attacks. The latency remains within an acceptable range, supporting the system’s feasibility for practical deployment in scenarios requiring semantic understanding and responsiveness.

%% file: sections/7_limitation.tex
\section{Limitations and Future Work}
\label{sec:limitations}

While VIM-Sense demonstrates strong performance in detecting visual information manipulation attacks in AR, several limitations remain. First, the system relies on VLMs operating in a zero-shot setting, which introduces inherent variability in responses. Although our use of structured chain-of-thought prompting improves stability, generative models may still yield inconsistent judgments, particularly in ambiguous cases. In response to this challenge, we will explore fine-tuning large-scale VLMs on AR-VIM or related datasets to improve their robustness in AR-specific semantic reasoning, especially as recent studies have demonstrated that even commercial models can be effectively fine-tuned using small-scale datasets~\cite{finetune01, finetune02}. Second, although VIM-Sense uses a lightweight edge-cloud architecture, the latency remains in the range of several seconds. \rb{This is suitable for general AR use cases, especially those that only require periodic attack detection and do not involve rapid changes in scene information and virtual content, such as indoor navigation, assembly training, and visual searching, but may be inadequate for high-speed or real-time use cases such as AR-assisted driving or human-robot collaboration.} We will monitor the advances in the generative AI field and explore techniques such as model compression and knowledge distillation, particularly for open-source VLMs, to integrate more lightweight alternatives when commercial APIs are impractical. These efforts will continuously incorporate faster and more efficient models into our system.


Beyond model performance, future improvements can be made by expanding the scope of our dataset.
\rb{Although AR-VIM covers a wide range of scenarios, its scope can be further expanded by incorporating additional data and involving user studies with more users. Currently, the virtual content in each scenario is manually designed, which is time-consuming. In future work, we will introduce automation into this process by enabling automated generation of virtual content, leveraging 3D content synthesis \cite{3dgeneration01, 3dgeneration02, 3dgeneration03} and placement\cite{xair, octo} methods. To ensure visual coherence, style transfer\cite{styletransfer01, styletransfer02} methods may also be applied to help virtual content blend seamlessly into the surrounding scene. On the user-based data validation side, participants currently label the videos with the knowledge that virtual content has been injected. In future work, we will conduct in-situ user studies in which users interact with AR content through the AR view alone, without being explicitly informed of any manipulation. This more immersive setting will reduce user awareness of potential attacks, making them more susceptible to attacks, and thereby enabling a more valid evaluation of the dataset.}

\rb{Meanwhile, the attack taxonomy of AR-VIM was mainly focused on the semantic information change, which can be further enriched by involving other types of attacks in AR. For example, cognitive overload\cite{congitiveoverload01, congitiveoverload02}, which introduces excessive or cluttered virtual elements and cause users to overlook critical real-world cues. Our future work will also consider risks introduced by non-adversarial AR content, where benign overlays may unintentionally trigger the same mechanisms of information obfuscation or introduce extra wrong information. Such content can lead to similar detrimental effects on user perception and decision-making, even though it is not intended to be malicious. Some samples in AR-VIM reflected this kind of effect, yet a more systematic data collection is still required.} Also, AR-VIM does not yet cover multi-modal attacks that may involve additional sensory channels such as audio\cite{audio02, audio01} or haptics\cite{haptic01, haptic02}, which are becoming increasingly common in real-world AR applications and may jointly influence user perception. Thus, we will explore multimodal cues to create more attack cases and extend the detection framework to handle such cross-modal attacks.

Finally, while VIM-Sense currently relies heavily on VLMs as the core reasoning engine, we will move beyond prompt engineering by incorporating other lightweight interpretable modules to assist the VLM. \rb{This will also enable VIM-Sense to produce more informative outputs, such as identifying the specific spatial regions affected by the attack, rather than providing only a binary decision.} One direction is to integrate multi-modal object detection models for visual grounding\cite{vg01, vg02}, enabling the system to verify whether the VLM’s reasoning corresponds to actual content in the AR scene. We will also introduce a form of context memory that tracks scene context or prior detections to support more consistent and context-aware prompt construction\cite{prompting01}. These extensions will make the reasoning process more robust, reduce reliance on individual prompts, and improve interpretability in real-world AR deployments.

%% file: sections/8_conclusion.tex
\section{Conclusion}
\label{sec:conclusion}

\begin{figure}[t]
\includegraphics[width=1.0\linewidth]{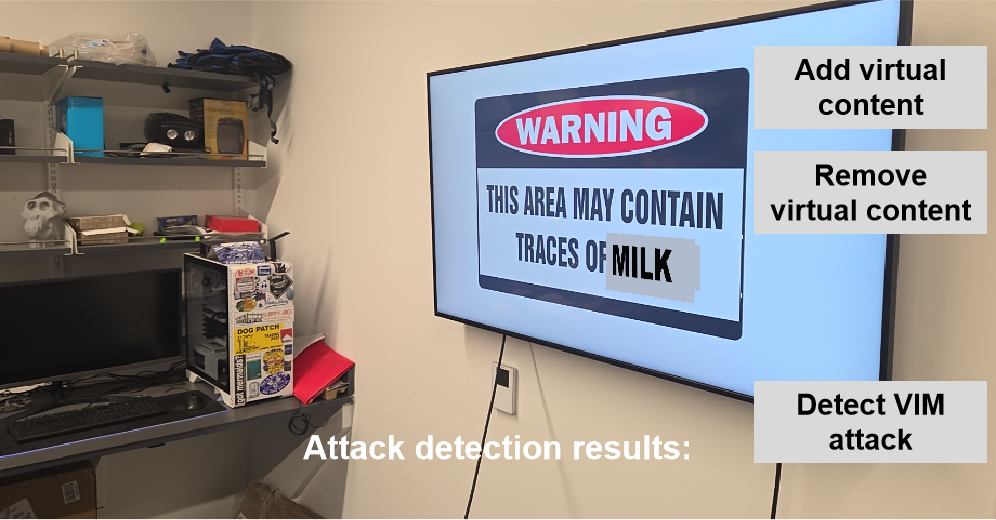}
\centering
\vspace{-0.35cm}
\caption{Screenshot of the real-world AR app implementation.}
\label{fig:real world ui}
\vspace{-0.65cm}
\end{figure}
 
In this work, we proposed a multimodal semantic reasoning approach to address the challenge of detecting visual information manipulation attacks in AR environments. We first proposed a mathematical formulation for such attacks, defining them based on attack formats and attack purposes. Building on this framework, we introduced the AR-VIM dataset, which consists of \rb{452} raw-AR video pairs across \rb{202} scenes, covers the defined attack types, and is validated through an IRB-approved user-based data labeling task. To detect the attacks, we developed VIM-Sense, which leverages VLMs and OCR-based prompt construction within an edge-cloud architecture. Our experiments show that VIM-Sense largely outperforms baseline methods in attack detection accuracy and achieves a moderate responsiveness. Finally, VIM-Sense is also deployed and tested with a real-world mobile AR application, demonstrating potential for safeguarding semantic integrity in AR applications.